\documentclass[10pt,twocolumn,letterpaper]{article}

\usepackage{cvpr}
\usepackage{times}
\usepackage{epsfig}
\usepackage{graphicx}
\usepackage{amsmath}
\usepackage{amssymb}
\usepackage{subcaption}
\usepackage{authblk}

\usepackage[pagebackref=true,breaklinks=true,letterpaper=true,colorlinks,bookmarks=false]{hyperref}

\cvprfinalcopy 

\def\cvprPaperID{7576} 

\ifcvprfinal\fi
\begin{document}

\title{MatrixNets: A New Scale and Aspect Ratio Aware Architecture \\ for Object Detection
}

\author[1,2]{Abdullah Rashwan }
\author[1]{Rishav Agarwal}
\author[1,3]{ Agastya Kalra }
\author[1,2]{Pascal Poupart } 
\affil[1]{University of Waterloo}
\affil[2]{Vector Institute}
\affil[3]{Akasha Imaging Corp}
\affil[ ]{\tt \small {\{arashwan,rragrawal,ppoupart\}@uwaterloo.ca}, {\tt \small agastya@akasha.im}}


\maketitle

\begin{abstract}
We present MatrixNets ($x$Nets), a new deep architecture for object detection. $x$Nets map objects with similar sizes and aspect ratios into many specialized layers, allowing $x$Nets to provide a scale and aspect ratio aware architecture. We leverage $x$Nets to enhance single-stage object detection frameworks. First, we apply $xNets$ on anchor-based object detection, for which we predict object centers and regress the top-left and bottom-right corners. Second, we use MatrixNets for corner-based object detection by predicting top-left and bottom-right corners. Each corner predicts the center location of the object. We also enhance corner-based detection by replacing the embedding layer with center regression. Our final architecture achieves mAP of 47.8 on MS COCO, which is higher than its CornerNet \cite{law2019cornernet} counterpart by +5.6 mAP while also closing the gap between single-stage and two-stage detectors. The code is available at \url{https://github.com/arashwan/matrixnet}.
\end{abstract}

\section{Introduction}

\begin{figure}[t]
\begin{center}
\includegraphics[width=0.8\linewidth]{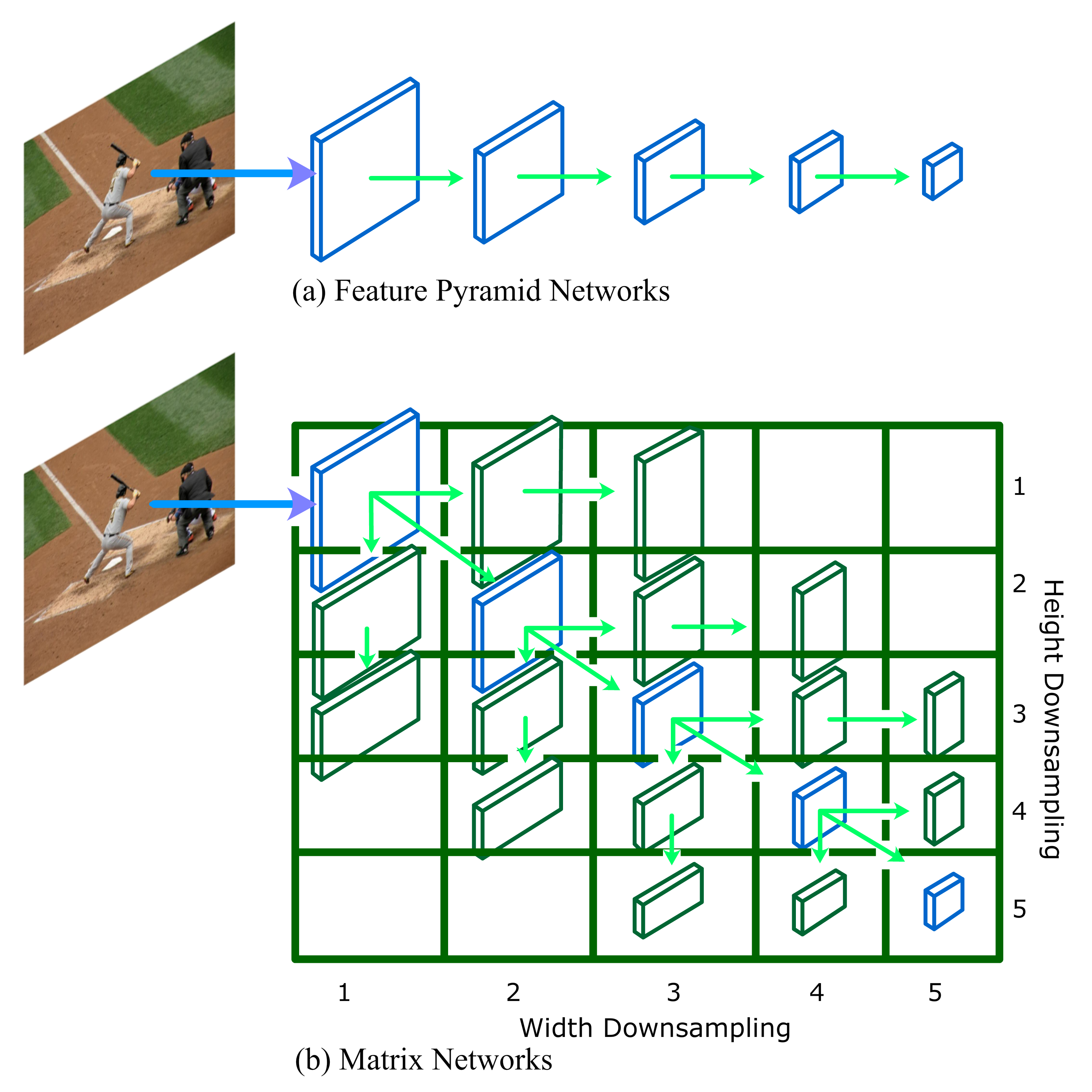}
\end{center}
\vspace{-0.25cm}
   \caption{(a) Shows the FPN architecture \cite{lin2017feature}, where there are different output layers assigned at each scale. Note we do not show the skip connections for the sake of simplicity. (b) Shows the MatrixNet architecture, where the 5 FPN layers are viewed as the diagonal layers in the matrix. We fill in the rest of the matrix by downsampling these layers.}
\label{fig:xnets}
\end{figure}

Object detection is one of the most widely studied tasks in computer vision with many applications to tasks such as object tracking, instance segmentation, and image captioning. Object detection architectures can be grouped into two categories: two-stage detectors \cite{li2019scale}, and one-stage detectors \cite{lin2017focal, law2018cornernet}. Two-stage detectors leverage a region proposal network to find a fixed number of object candidates. Then a second network is used to predict a score for each candidate and to refine its bounding box. Furthermore, one-stage detectors can also be split into two categories: anchor-based detectors \cite{lin2017focal, zhu2019feature} and corner (or key-points) based detectors \cite{law2018cornernet, duan2019centernet}. Anchor-based detectors contain many anchor boxes, and they predict offsets and classes for each anchor. On the other hand, corner based detectors \cite{law2018cornernet,duan2019centernet} predict top-left and bottom-right corner heat-maps and match them together using feature embeddings.



\begin{figure}[t]
\begin{center}
\includegraphics[width=0.85\linewidth]{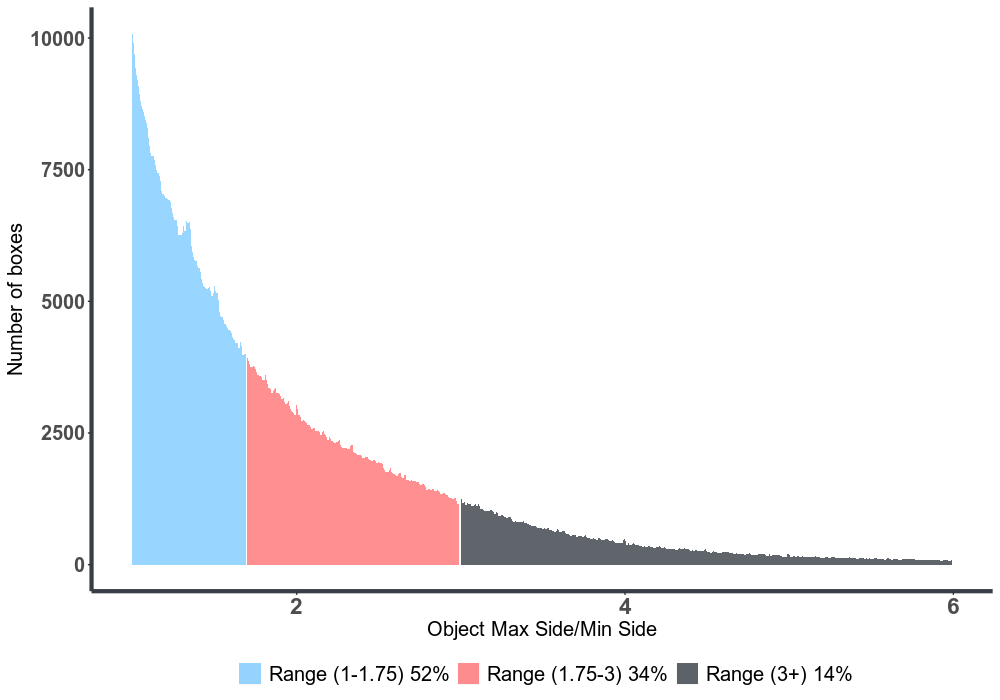}
\vspace{-0.25cm}
\end{center}
  \caption{Histogram of number of boxes vs the ratio of maximum dimension to minimum dimension of the object.}
\label{fig:long}
\label{fig:histogram}
\vspace{-0.5cm}
\end{figure}

Detecting objects at different scales is a major challenge for object detection. One of the biggest advancements in scale aware architectures was Feature Pyramid Networks (FPNs) \cite{lin2017feature}. FPNs were designed to be scale-invariant by having multiple layers with different receptive fields so that objects are mapped to layers with relevant receptive fields. Small objects are mapped to earlier layers in the pyramid, and larger objects are mapped to later layers. Since the size of the objects relative to the downsampling of the layer is kept nearly uniform across pyramid layers, a single output sub-network can be shared across all layers. Although FPNs provided an elegant way for handling objects of different sizes, they did not provide any solution for objects of different aspect ratios. Objects such as a high tower, a giraffe, or a knife introduce a design difficulty for FPNs: does one map these objects to layers according to their width or height? Assigning the object to a layer according to its larger dimension would result in loss of information along the smaller dimension due to aggressive downsampling, and vice versa. This problem is prevalent in datasets like MS-COCO \cite{lin2014microsoftcoco}. Fig. \ref{fig:histogram} shows the histogram of the number of objects versus the values of the maximum side of an object divided by the minimum side. We found that ~50\% of the objects have max/min values higher than 1.75, and ~14\% have max/min values greater than 3. Hence, modelling these rectangular objects efficiently is essential for good detection performance.
In this work, we introduce MatrixNets ($x$Nets), a new scale and aspect ratio aware CNN architecture. $x$Nets, as shown in Fig.~\ref{fig:xnets}, have several matrix layers, each layer handles an object of specific size and aspect ratio. $x$Nets assign objects of different sizes and aspect ratios to layers such that object sizes within their assigned layers are close to uniform. This assignment allows a square output convolution kernel to equally gather information about objects of all aspect ratios and scales. $x$Nets can be applied to any backbone, similar to FPNs. We denote this by appending a "-X" to the backbone, i.e. ResNet50-X \cite{he2016deep}. As an application for $x$Nets, we first use $x$Nets for anchor-based one-stage object detection. Instead of using multiple anchor boxes per feature map, we decided to consider the case where there is only one box per feature map, making it similar to an anchor free architecture. In a second application, we use $x$Net for corner-based object detection. We show how to leverage $x$Net to improve the CornerNet architecture. On MS-COCO, we set the new state-of-the-art performance (47.8 mAP) for human-designed single-stage detectors.

The rest of the paper is structured as follows. Section \ref{sec:related_work} reviews related work. Section \ref{sec:MatrixNets} formalizes the idea of MatrixNets. Section \ref{sec:MatrixNetsApplications} discusses the application of MatrixNets to different mainstream single-stage object detection frameworks. Section \ref{sec:Experiments} covers the experiments, including a thorough ablation analysis. We conclude the paper in Section \ref{sec:Conclusions}.

\section{Related Work}
\label{sec:related_work}

\textbf{Two-Stage Detectors:} Two-stage detectors generate the final detection by first extracting RoIs, then in a second stage, hence the name, classifying and regressing on top of each RoIs. The two-stage object detection paradigm was first introduced by R-CNN \cite{girshick2014richrcnn}. R-CNN used the selective search method \cite{uijlings2013selective} to propose RoIs, then a CNN network is used to score and refine the RoIs. Fast-RCNN \cite{girshick2015fast} and SPP \cite{he2015spatial} improved R-CNN by extracting RoIs from feature maps rather than the input image. Faster-RCNN \cite{ren2015faster} introduced the Region Proposal Network (RPN), which is a trainable CNN that generates the RoIs allowing the two-stage detectors to be trained end-to-end. Several improvements to the Faster-RCNN framework have been proposed since \cite{cai2018cascade,lin2017feature,li2019scale}.

\textbf{One-Stage Detectors:} Anchor-based detection is the most common framework for single-stage object detectors. Anchor-based detectors generate the detections by directly classifying and regressing the pre-defined anchors. One of the first single-stage detectors, YOLO \cite{redmon2016you,redmon2017yolo9000}, is still widely used since it can be run in real time. One-stage detectors tend to be superior in speed, but lagging in performance when compared to two-stage detectors. RetinaNet \cite{lin2017focal} was the first attempt to close the gap between the two paradigms. RetinaNet proposed the focal loss to help correct for the class imbalance of positive to negative anchor boxes. RetinaNet uses a hand-crafted heuristic to assign anchors to ground-truth objects using Intersection-Over-Union (IOU). Recently, it has been found that improving the anchors to ground-truth object assignments can have a significant impact on the performance \cite{zhu2019feature, zhang2019freeanchor}. Further, Feature Selective Anchor-Free (FSAF) \cite{zhu2019feature} ensembles the anchor-based output with an anchor free output head to improve performance. AnchorFree \cite{zhang2019freeanchor} improved the anchor to the ground-truth matching process by formulating the problem as a maximum likelihood estimation (MLE). 

Another framework for one-stage detection is corner-based (or keypoint-based) detectors which was first introduced by CornerNet \cite{law2018cornernet}. CornerNet predicts top-left and bottom-right corner heat-maps and match them together using feature embeddings. CenterNet \cite{duan2019centernet} substantially improved CornerNet architecture by predicting object centers along with corners.

\section{MatrixNets}
\label{sec:MatrixNets}

MatrixNets ($x$Nets), as shown in Fig. \ref{fig:xnets}, model objects of different sizes and aspect ratios using a matrix of layers where each entry $i,j$ in the matrix represents a layer, $l_{i,j}$. Each layer $l_{i,j}$ has a width down-sampling of $2^{i-1}$ and height down-sampling of $2^{j-1}$. The top left layer (base layer) is $l_{1,1}$ in the matrix.  The diagonal layers are square layers of different sizes, equivalent to an FPN, while the off-diagonal layers are rectangle layers, unique to $x$Nets. Layer $l_{1,1}$ is the largest layer in size, every step to the right cuts the width of the layer by half, while every step-down cuts the height by half. For example, $Width(l_{3,4}) =  0.5 Width(l_{3,3})$. Diagonal layers model objects with square-like aspect ratios, while off-diagonal layers model objects with more extreme aspect ratios. Layers close to the top right or bottom left corners of the matrix model objects with very high or very low aspect ratios. Such objects are scarce, so these layers can be pruned for efficiency.

\subsection{Layer Generation}

Generating matrix layers is a crucial step since it impacts the number of model parameters. The more parameters, the more expressive the model, but the harder the optimization problem. In our method, we chose to introduce as few new parameters as possible. The diagonal layers can be obtained from different stages of the backbone or using a feature pyramid backbone  \cite{lin2017feature}. The upper triangular layers are obtained by applying a series of shared 3x3 convolutions with stride 1x2 on the diagonal layers. Similarly, the left bottom layers are obtained using shared 3x3 convolutions with stride 2x1. This sharing helps reduce the number of additional parameters introduced by the matrix layers.

\subsection{Layer Ranges}
We define the range of widths and heights of objects assigned to each layer in the matrix to allow each layer to specialize. The ranges need to reflect the receptive field of the feature vectors of the matrix layers. Each step to the right in the matrix effectively doubles the receptive field in the horizontal dimension, and each step down doubles the receptive field in the vertical dimension. Hence, the range of the widths or heights needs to be doubled as we advance to the right or down in the matrix. Once the range for the first layer $l_{1,1}$ is defined, we can generate the ranges for the rest of the matrix layers using the above rule. For example, if the range for layer $l_{1,1}$ (base layer) is $H\in [24px,48px]$, $W\in [24px,48px]$, the range for layer $l_{1,2}$ will be $H\in [24,48]$, $W\in [48,96]$. We show multiple layer ranges in our ablation studies.

Objects on the boundaries of these ranges could destabilize training since layer assignment would change if there is a slight change in object size. To avoid this problem, we relax the layer boundaries by extending them in both directions. This relaxation is accomplished by multiplying the lower end of the range by a number less than one, and the higher end by a number greater than one. In all our experiments, we use 0.8, and 1.3 respectively.

\subsection{Advantages of MatrixNets}
The key advantage of MatrixNets is that they allow a square convolutional kernel to accurately gather information about different aspect ratios. In traditional object detection models, such as RetinaNet, a square convolutional kernel is required to output boxes of different aspect ratios and scales. Using a square convolutional kernel is counter-intuitive since boxes of different aspect ratios and scales require different contexts. In MatrixNets, the same square convolutional kernel can be used for detecting boxes of different scales and aspect ratios since the context changes in each matrix layer. Since object sizes are nearly uniform within their assigned layers, the dynamic range of the widths and heights is smaller compared to other architecture such as FPNs. Hence, regressing the heights and widths of objects becomes an easier optimization problem. Finally, MatrixNets can be used as a backbone to any object detection architecture, anchor-based or keypoint-based, one-stage or two-stage detectors.

\section{MatrixNets Applications}
\label{sec:MatrixNetsApplications}
In this section, we show that MatrixNets can be used as a backbone for two single-shot object detection frameworks; center-based and corner-based object detection. In center-based object detection, we predict the object centers while regressing the top-left and bottom-right corners. In corner-based object detection, we predict the object corners and regress the center of the object. Corners that predict the same center are matched together to form a detection.

\begin{figure}[t]
\begin{center}
\includegraphics[width=1.0\linewidth]{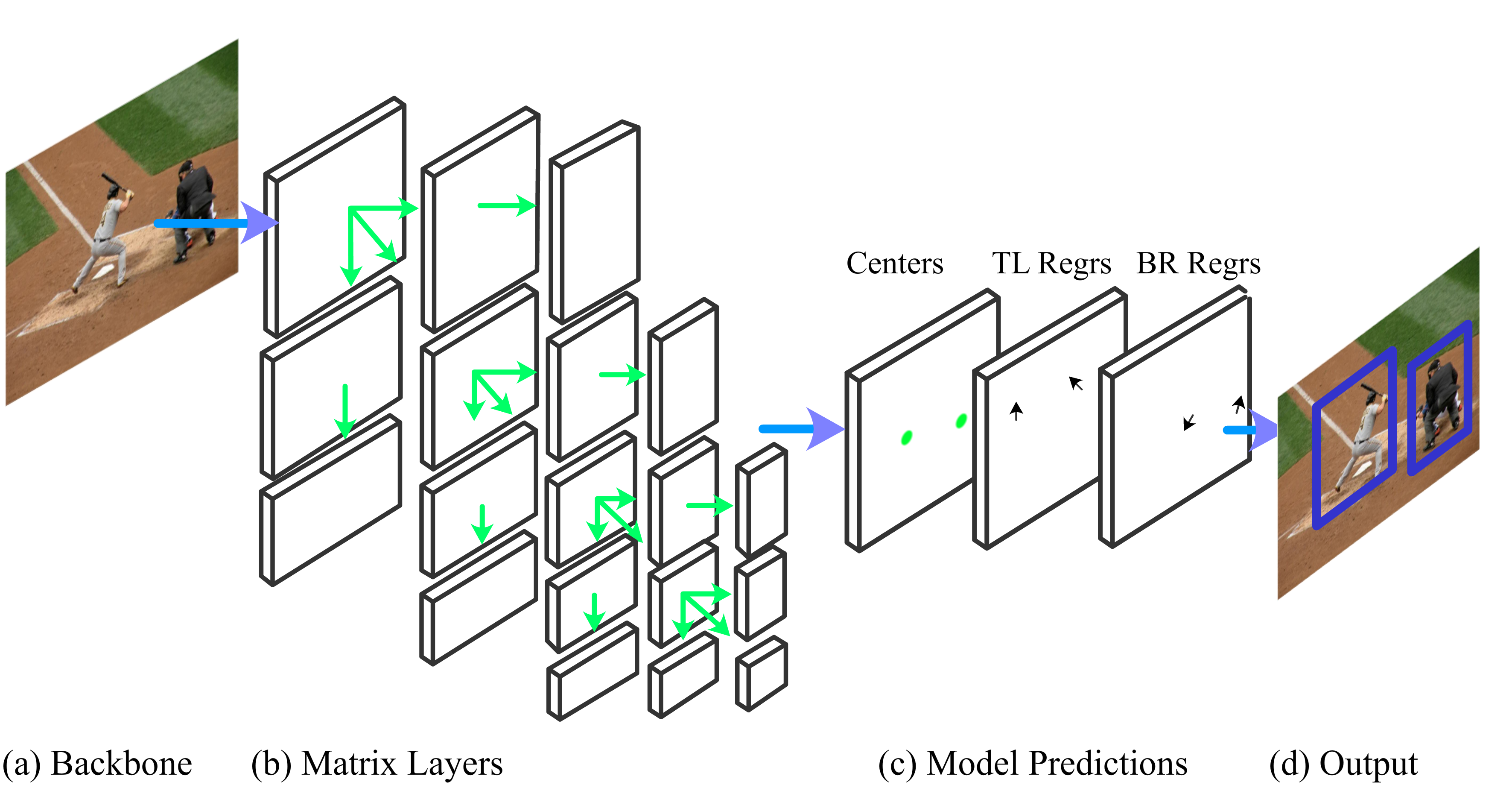}
\end{center}
  \caption{The Centers-$x$Net architecture.}
\label{fig:long}
\label{fig:AnxNet}
\end{figure}

\subsection{Center-based Object Detection}

Anchor-based object detection is a common framework for single-stage object detection. 
Using MatrixNet as a backbone naturally handles objects of different scales and aspect ratios. Although using multiple anchors of different scales can potentially improve the performance, we decided to simplify the architecture by using a single anchor per location making it anchor free. Hence, ground truth objects can be assigned to the nearest center location during the training.

\subsubsection{Center-based Object Detection Using MatrixNets}

As shown in Fig. \ref{fig:AnxNet}, our Centers-$x$Net architecture consists of 4 stages. (a-b) We use a $x$Net backbone as defined in Section \ref{sec:MatrixNets}. (c) Using a shared output sub-network, for each matrix layer, we predict the center heatmaps, top-left corner regressions, and bottom-right corner regressions for objects within their layers. (d) We combine the outputs of all layers with soft non-maximum suppression \cite{bodla2017softnms} to achieve the final output.

\textbf{Center Heatmaps}
During the training, ground truth objects are first assigned to layers in the matrix according to their widths and heights. Within the layer, objects are assigned to the nearest center location. To deal with unbalanced classes, we use focal loss \cite{lin2017focal}.

\textbf{Corner Regression}
Object sizes are bounded with the matrix layers, which makes it feasible to regress object top-left and bottom-right corners. As shown in Fig. \ref{fig:AnxNet}, for each center, Centers-$x$Net predicts the corresponding top-left and bottom-right corners. During the training, we use the smooth L1 loss for parameter optimization.

\textbf{Training}
We use a batch size of 23 for all experiments. During the training, we use crops of sizes 640x640, and we use a standard scale jitter of 0.6-1.5. For optimization, we use the Adam optimizer and set an initial learning rate to 5e-5, and cut it by 1/10 after 250k iterations, training for a total of 350k iterations. For our matrix layer ranges, we set $l_{1,1}$ to be [24px-48px]x[24px-48px] and then scale the rest as described in Section~\ref{sec:MatrixNets}.

\textbf{Inference}
For single-scale inference, we resize the max side of the image to 900px. We use the original and the horizontally flipped images as an input to the network. For each layer in the network, we choose the top 100 center detections. Corners are computed using the top-left and bottom-right corner regression outputs. The bounding boxes of the original image and the flipped ones are mixed. Soft-NMS \cite{bodla2017softnms} layer is used to reduce redundant detections. Finally, we choose the top 100 detections according to their scores as the final output of the detector.

\begin{figure}[t]
\begin{center}
\includegraphics[width=1.0\linewidth]{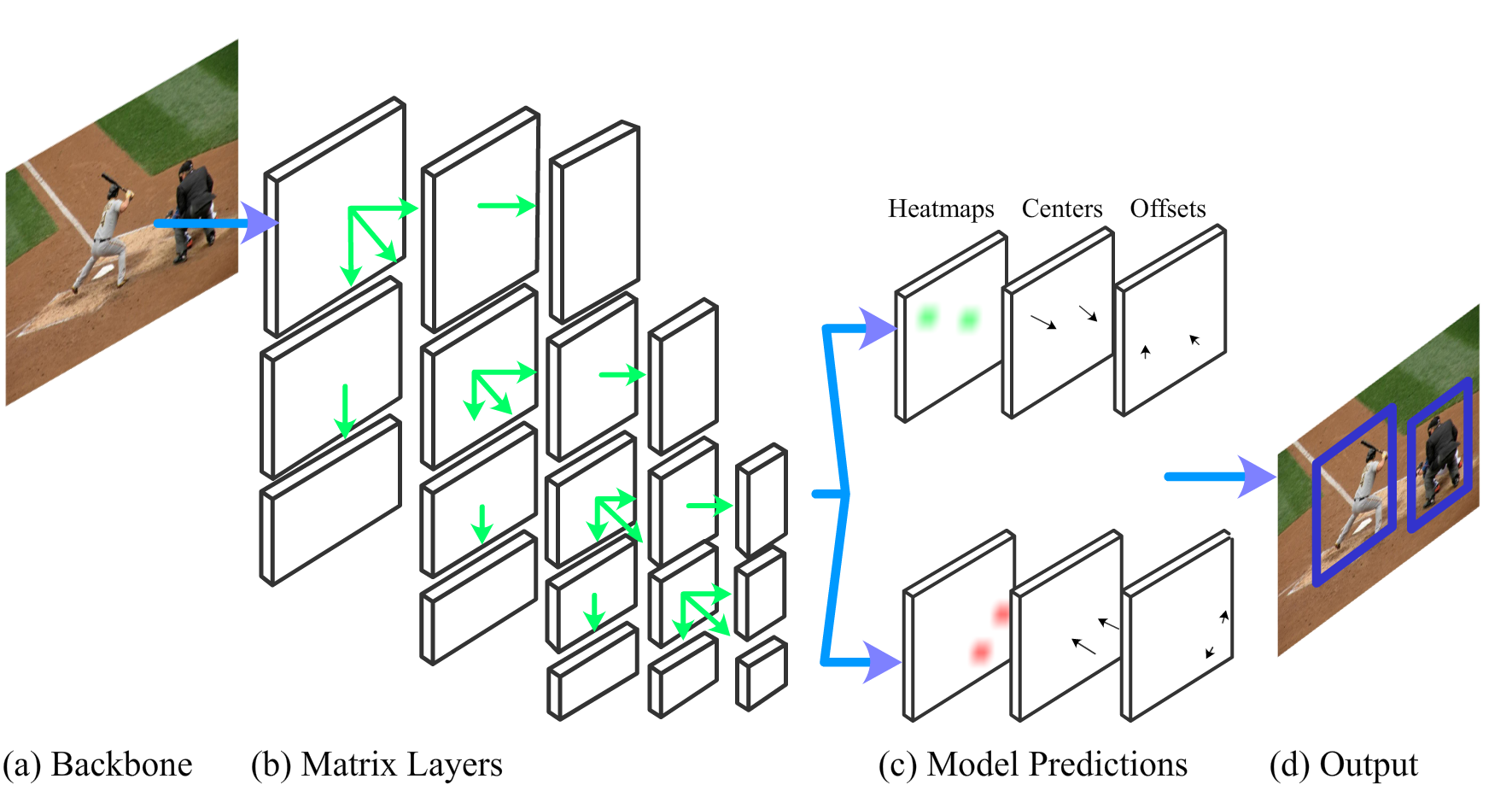}
\end{center}
  \caption{The Corners-$x$Net architecture.}
\label{fig:long}
\label{fig:KPxNet}
\end{figure}

\subsection{Corner-based Objection Detection}
CornerNet \cite{law2018cornernet} was proposed as an alternative to anchor-based detectors, CornerNet predicts a bounding box as a pair of corners: top-left, and bottom-right. For each corner, CornerNet predicts heatmaps, offsets, and embeddings. Top-left and bottom-right corner candidates are extracted from the heatmaps. Embeddings are used to group the top-left, and bottom-right corners that belong to the same object. Finally, offsets are used to refine the bounding boxes producing tighter bounding boxes. This approach has three main limitations. 
\begin{enumerate}

\item CornerNet handles objects from different sizes and aspect ratios using a single output layer. As a result, predicting corners for large objects presents a challenge since the available information about the object at the corner location is not always available with regular convolutions. To solve this challenge, CornerNet introduced the corner pooling layer that uses a max operation on the horizontal and vertical dimensions. The top left corner pooling layer scans the entire right bottom image to detect any presence of a corner. Eventhough experimentally, it is shown that corner pooling stabilizes the model, we know that max operations lose information. For example, if two objects share the same location for the top edge, only the object with the max features will contribute to the gradient. So, we can expect to see false positive predictions due to corner pooling layers.

\item Matching the top left and bottom right corners is done with feature embeddings. Two problems arise from using embeddings in this setting. First, the pairwise distances need to be optimized during the training, so as the number of objects in an image increases, the number of pairs increases quadratically, which affects the scalability of the training when dealing with dense object detection. The second problem is learning embeddings themselves. CornerNet tries to learn the embedding for each object corner conditioned on the appearance of the other corner of the object. Now, if the object is too big, the appearance of both corners can be very different due to the distance between them. As a result, the embeddings at each corner can be different, as well. Also, if there are multiple objects in the image with a similar appearance, the embeddings for their corners will likely be similar. This is why we saw examples where CornerNet merged persons or traffic lights.
\vspace{-0.25cm}

\item As a result of the previous two problems, CornerNet is forced to use the Hourglass-104 backbone to achieve state-of-the-art performance. Hourglass-104 has over 200M parameters, very slow and unstable training, requiring 10 GPUs with 12GB memory to ensure a large enough batch size for stable convergence.

\vspace{-0.25cm}
\end{enumerate}{}

\subsubsection{Corner-based Object Detection Using MatrixNets}
Fig.~\ref{fig:KPxNet} shows our proposed architecture for corner-based object detection, Corners-$x$Net. Corners-$x$Net consists of 4 stages. (a-b) We use a $x$Net backbone as defined in Section 2. (c) Using a shared output sub-network, for each matrix layer, we predict the top-left and bottom-right corner heatmaps, corner offsets, and center predictions for objects within their layers. (d) We match corners within the same layer using the center predictions and then combine the outputs of all layers with soft non-maximum suppression to achieve the final output.

\textbf{Corner Heatmaps}
Using $x$Nets ensures that the context required for objects within a layer is bounded by the receptive field of a single feature map in that layer. As a result, corner pooling is no longer needed; regular convolutional layers can be used to predict the heatmaps for the top left and bottom right corners. Similar to CornerNet, we use the focal loss to deal with unbalanced classes.

\textbf{Corner Regression}
Due to image downsampling, refining the corners is important to have tighter bounding boxes. When scaling down a corner to $x$, $y$ location in a layer, we predict the offsets so that we can scale up the corner to the original image size without losing precision. We keep the offset values between $-0.5$, and $0.5$, and we use the smooth L1 loss to optimize the parameters.

\begin{table*}[h!]
\centering
\footnotesize
\caption{Performance comparison between Centers-$x$Net and Corners-$x$Net.}
\begin{tabular}{l|c|c|lll|lll}
\textbf{Backbone}  & \textbf{Test Image Size} (px) & \textbf{Inference Times} (ms) & \textbf{$AP$}      & \textbf{$AP_{50}$} & \textbf{$AP_{75}$} & \textbf{$AP_{S}$} & \textbf{$AP_{M}$} & \textbf{$AP_{L}$} \\
\hline

 & 800 & 247 & 42.7 & 61.9  & 46.6 & 22.6 & 48.0 & 59.2\\
Resnet-152-X-Centers & 900 & 265 & 43.6 & 62.3 & 47.5 & 24.0 & 48.4 & 59.1\\
 & 1000 & 275 & 44.0 & 63.0 & 48.2 & 25.0 & 49.0 & 57.9 \\
\hline \hline
 & 800 & 340 & 44.6 & 63.4 & 48.8 & 24.6 & 49.5 & 61.0 \\
Resnet-152-X-Corners & 900 & 355 & 44.7 & 63.6 & 48.7 & 26.0 & 49.2 & 59.8 \\
 & 1000 & 385 & 44.5 & 63.5 & 48.7 & 27.7 & 48.9 & 58.0\\
\end{tabular}
\label{table:cornersvscenters}
\end{table*}

\begin{figure*}[ht!]
\setlength\tabcolsep{1.5pt}
\begin{tabular}{cccccc}
 \includegraphics[height=25mm]{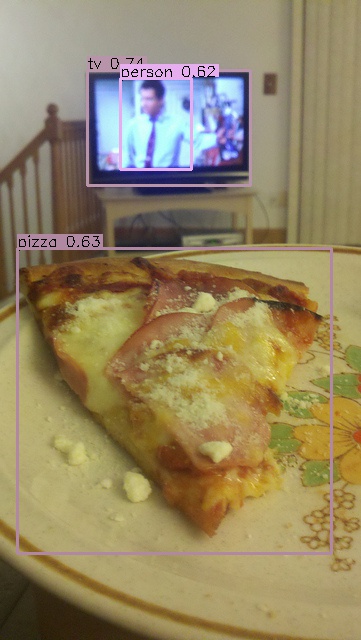}  \includegraphics[height=25mm]{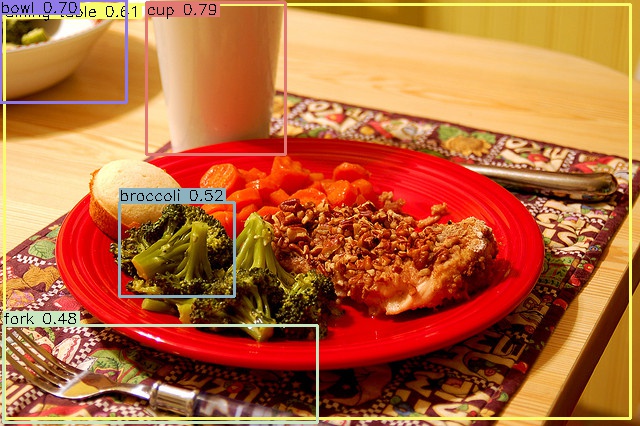} & \includegraphics[height=25mm]{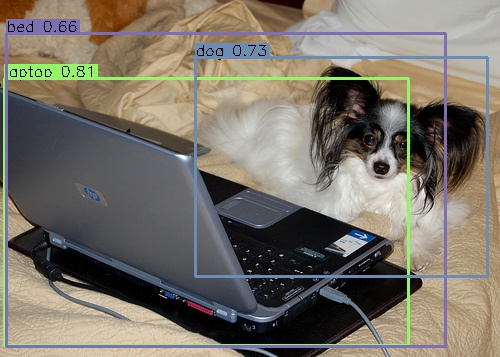}  
  \includegraphics[height=25mm]{figures/centers_c2.jpg} &
  \includegraphics[height=25mm]{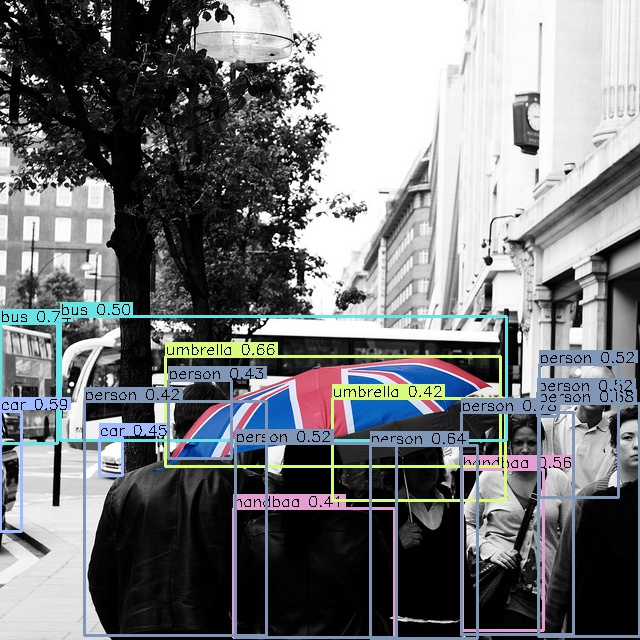}   \includegraphics[height=25mm]{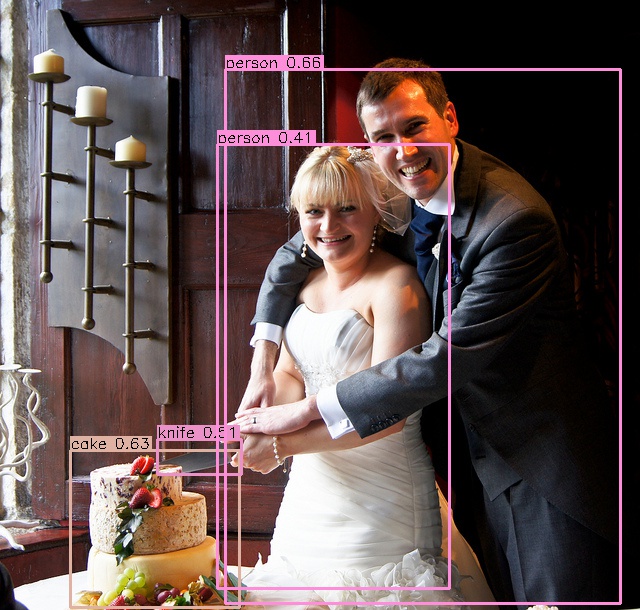} \\
(a) & (b) &  (c)  \\[6pt]
 \includegraphics[height=25mm]{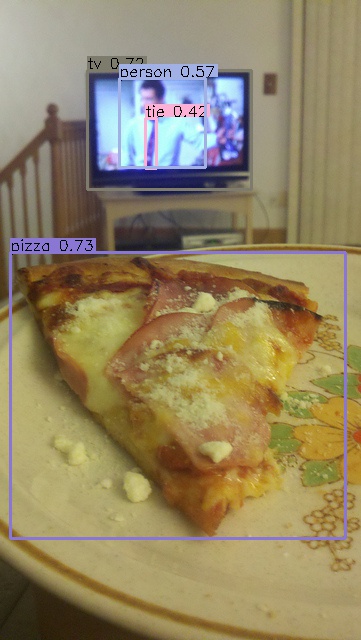}  \includegraphics[height=25mm]{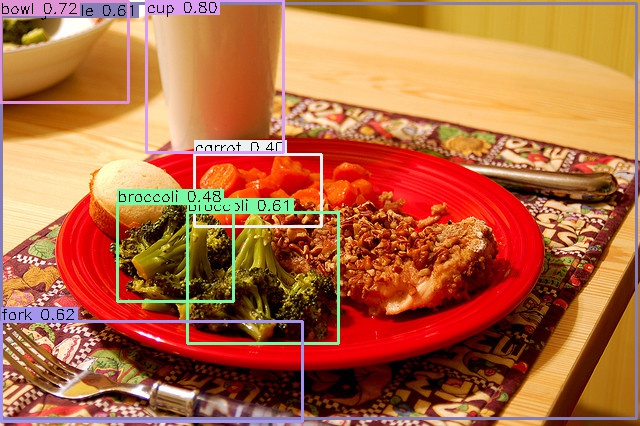} & \includegraphics[height=25mm]{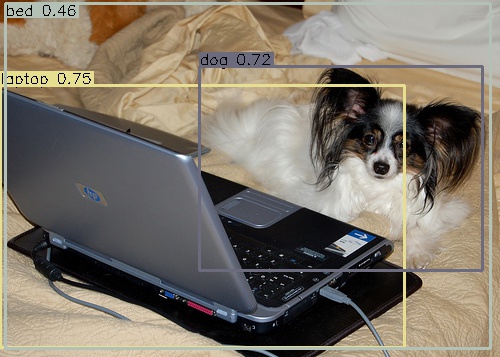}  
  \includegraphics[height=25mm]{figures/corners_c2.jpg} &
   \includegraphics[height=25mm]{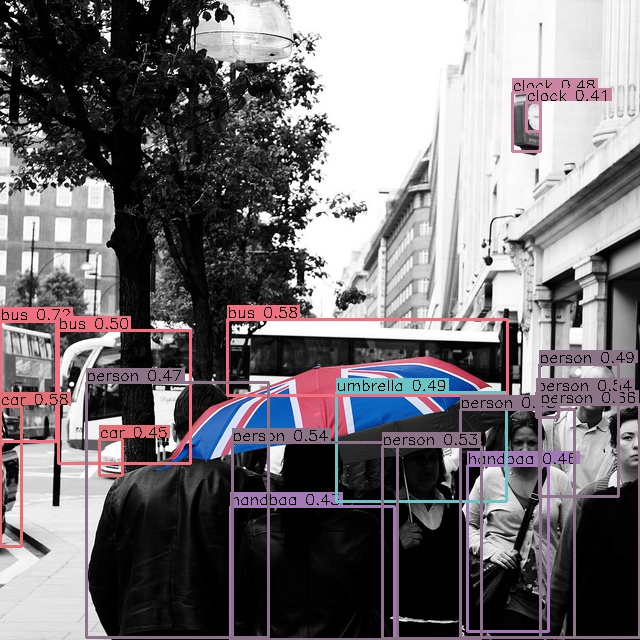}    \includegraphics[height=25mm]{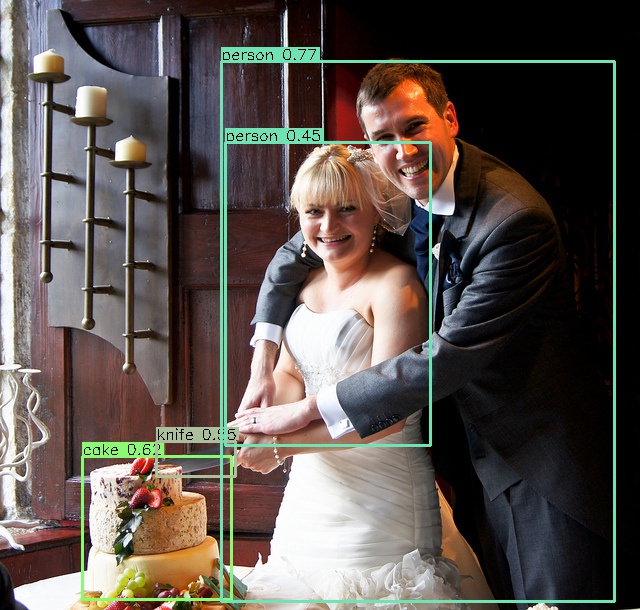} \\
(a) & (b) &  (c)  \\[6pt]
\end{tabular}
\vspace{-0.25cm}
\caption{Sample detection results for center-based object detector (top), and corner-based object detector (bottom) when resnet152-X is used as a backbone. Any detection with probability less than 0.4 is removed.}
\label{fig:centervscorner}
\vspace{-0.25cm}
\end{figure*}
\textbf{Center Regression}
Since the matching is done within each matrix layer,  the width and height of the object are guaranteed to be within a certain range. The center of the object can be regressed easily because the range for the centers is small. In CornerNet, the dynamic range for the centers is large, trying to regress centers in a single output layer would probably fail. Once the centers are obtained, the corners can be matched together by comparing the regressed centers to the actual center between the two corners. During the training, center regression scales linearly with the number of objects in the image compared to quadratic growth in the case of learning embeddings. To optimize the parameters, we use smooth L1 loss.

\textbf{Corners Matching}
For any pair of corners, the correct center is the average of their x and y locations. The relative distance between the correct center and each corner is the correct values for center regression for both corners if they belong to the same object. Hence, if both corners predict the center with an error rate of $30\%$ or lower, we match these corners together.

\textbf{Training}
We use a batch size of 23 for all experiments. During the training, we use crops of sizes 512x512, and we use a standard scale jitter of 0.6-1.5. For optimization, we use the Adam optimizer and set an initial learning rate of 5e-5, and cut it by 1/10 after 250k iterations, training for a total of 350k iterations. For our matrix layer ranges, we set $l_{1,1}$ to be [24px-48px]x[24px-48px] and then scale the rest as described in Section \ref{sec:MatrixNets}.

\textbf{Inference}
For single-scale inference, we resize the max side of the image to 900px. We use the original and the horizontally flipped images as an input to the network. For each layer in the network, we choose the top 50 top-left and bottom-right corners. The corner location is refined using corner regression outputs. Then, each pair of corners are matched together, as we discussed above. The bounding boxes of the original image and the flipped ones are mixed. Soft-nms \cite{bodla2017softnms} layer is used to reduce redundant detections. Finally, we choose the top 100 detections according to their scores as the final output of the detector.

Corners-$x$Net solves the problem (1) of CornerNets because all the matrix layers represent different scales and aspect ratios rather than having them all in a single layer. This also allows us to get rid of the corner pooling operation. (2) is solved since we no longer predict embeddings. Instead, we regress centers directly. By solving the first two problems of CornerNets, we will show in the experiments that we can achieve significantly higher results than CornerNet.

\section{Experiments}
\label{sec:Experiments}

We train all of our networks on a server with 8 Titan XP GPUs. Our implementation is done in PyTorch \cite{paszke2017automatic}, and the code will be made publicly available. For evaluation, we used the MS COCO detection dataset \cite{lin2014microsoftcoco}. We trained our models on MS COCO 'train-2017' set, validated on 'val-2017' and tested on the 'test-dev2017' set. For comparisons between our models, and ablation study, we reported the numbers on the 'val-2017' set. For our comparison to other detectors, we reported the numbers on 'test-dev2017'.

In the following subsections, we make a comparison between the performance of the Centers-$x$Net and Corners-$x$Net detectors. Then, we compare our detectors to other detectors. Finally, we show an ablation study through a set of experiments for evaluating different parts of the models.

\subsection{Centers-$x$Net vs Corners-$x$Net}
In this experiment, we wanted to compare the performance of Centers-$x$Net and Corners-$x$Net. As far as we know, this is the first fair comparison between both frameworks since both are sharing the same backbone (Resnet-152-X), training and inference settings.

Table \ref{table:cornersvscenters} shows the performance of both architectures at different test image sizes. mAP numbers are reported on MS COCO 'val-2017' set. Centers-$x$Net performs the best at a test image size of $1000$px, while Corners-$x$Net performs the best at a test image size of $900$px. Overall, Corners-$x$Net performs better than Centers-$x$Net in terms of mAP numbers. 
Corners-$x$Net seems more robust to varying image sizes, and the mAP drops by $<$0.2 mAP when varying the test image size by $\pm 100$px. On the other hand, Centers-$x$Net is very sensitive to test image size, and there is a performance drop of $<$1.3 mAP when varying the test image size by $\pm 100$px. The reason for such a drop in performance in the case of Centers-$x$Net is that an object can be entirely missed if two centers of the same objects collide at the same location. Since Centers-$x$Net is equivalent to using one anchor per location, the probability of collisions increases as the test image size decreases. Hence, we can see a drop in performance as we decrease the test image size. For Corners-$x$Net, an object can be missed if collisions happen at both corners, which has a much lower probability compared to the Centers-$x$Net case.

As per inference time, Centers-$x$Net architecture is ~$100$ms faster than Corners-$x$Net at all test image sizes. Corners-$x$Net uses more prediction outputs than Centers-$x$Net. Also, corner matching uses GPU and CPU time. Hence, there is an overhead of ~$100$ms using Corners-$x$Net.

One more observation from Table \ref{table:cornersvscenters}, test image sizes directly impact AP for small, medium, and large objects. This observation can be used to tune the mAP on the set of objects we are chiefly interested in. For consistency, and from this point until the end of this paper, we fix the test image sizes at $900$px.

\begin{table*}[ht!]
\centering
\caption{State-of-the-art comparison on COCO test-dev2017 set.
A blank indicates methods for which results were not available }
\resizebox{\textwidth}{!}{
\begin{tabular}{l|l|rrrrrr|rrrrrr} 
\hline
Architecture & Backbone & \multicolumn{1}{l}{mAP} & \multicolumn{1}{l}{$AP_{50}$} & \multicolumn{1}{l}{$AP_{75}$} & \multicolumn{1}{l}{$AP_S$} & \multicolumn{1}{l}{$AP_M$} & \multicolumn{1}{l}{$AP_L$} & \multicolumn{1}{|l}{$AR_1$} & \multicolumn{1}{l}{$AR_{10}$} & \multicolumn{1}{l}{$AR_{100}$} & \multicolumn{1}{l}{$AR_S$} & \multicolumn{1}{l}{$AR_M$} & \multicolumn{1}{l}{$AR_L$}  \\ 
\hline
RetinaNet \cite{lin2017focal} & ResNeXt 101 \cite{xie2017aggregated}  & 40.8 & 61.1 & 44.1 & 24.1 & 44.1 & 51.2 & \multicolumn{1}{l}{} & \multicolumn{1}{l}{} & \multicolumn{1}{l}{} & \multicolumn{1}{l}{} & \multicolumn{1}{l}{} & \multicolumn{1}{l}{} \\
FCOS \cite{tian2019fcos} & ResNeXt 101 & 42.1 & 62.2 & 46.1 & 26.0 & 45.6 & 52.6 & \multicolumn{1}{l}{} & \multicolumn{1}{l}{} & \multicolumn{1}{l}{} & \multicolumn{1}{l}{} & \multicolumn{1}{l}{} & \multicolumn{1}{l}{} \\
FSAF \cite{zhu2019feature} & ResNeXt 101 & 42.9 & 63.8 & 46.3 & 26.6 & 46.2 & 52.7 & \multicolumn{1}{l}{} & \multicolumn{1}{l}{} & \multicolumn{1}{l}{} & 
\multicolumn{1}{l}{} & \multicolumn{1}{l}{} & \multicolumn{1}{l}{}\\
FSAF (Multi-Scale) \cite{zhu2019feature}   & ResNeXt 101 & 44.6 & 65.2 & 48.6 & 29.7 & 47.1 & 54.6 & \multicolumn{1}{l}{} & \multicolumn{1}{l}{} & \multicolumn{1}{l}{}       & \multicolumn{1}{l}{} & \multicolumn{1}{l}{} & \multicolumn{1}{l}{} \\ 
Free-Anchor \cite{zhang2019freeanchor} & ResNeXt 101 & 44.8 & 64.3 & 48.4 & 27.0 & 47.9 & 56.0 & \multicolumn{1}{l}{} & \multicolumn{1}{l}{} & \multicolumn{1}{l}{} & 
\multicolumn{1}{l}{} & \multicolumn{1}{l}{} & \multicolumn{1}{l}{} \\
\hline
CornerNet    \cite{law2018cornernet} & Hourglass-54  & 37.8 & 53.7 & 40.1 & 17.0 & 39.0 & 50.5 & 33.9 & 52.3 & 57.0 & 35 & 59.3 & 74.7\\
CornerNet (Multi-Scale) \cite{law2018cornernet}  & Hourglass-54  & 39.4 & 54.9 & 42.3 & 18.9 & 41.2 & 52.7 & 35.0 & 53.5 & 57.7 & 36.1 & 60.1 & 75.1 \\

CenterNet \cite{duan2019centernet} & Hourglass-54  & 41.6 & 59.4  & 44.2 & 22.5  & 43.1 & 54.1 & 34.8  & 55.7 & 60.1 & 38.6 & 63.3 & 76.9\\
CornerNet-Lite  \cite{law2019cornernet} & Hourglass-54* & 43.2 & \multicolumn{1}{l}{}      & \multicolumn{1}{l}{}      & 24.4                     & 44.6                    & 57.3                     & \multicolumn{1}{l}{}     & \multicolumn{1}{l}{}      & \multicolumn{1}{l}{}       & \multicolumn{1}{l}{}     & \multicolumn{1}{l}{}     & \multicolumn{1}{l}{}      \\

CenterNet (Multi-Scale)\cite{duan2019centernet}        & Hourglass-54  & 43.5  & 61.3 & 46.7 & 25.3 & 45.3 & 55.0 & 36.0 & 57.2 & 61.3 & 41.4 & 64.0 & 76.3\\
\hline
CornerNet   \cite{law2018cornernet}                     & Hourglass 104 & 40.5 & 56.5 & 43.1 & 19.4 & 42.7 & 53.9 & 35.3 & 54.3 & 59.1 & 37.4 & 61.9 & 76.9 \\
CornerNet (Multi-Scale)   \cite{law2018cornernet}       & Hourglass 104 & 42.1 & 57.8 & 45.3 & 20.8 & 44.8 & 56.7 & 36.4  & 55.7  & 60.0 & 38.5 & 62.7 & 77.4 \\
CenterNet \cite{duan2019centernet}                       & Hourglass 104 & 44.9 & 62.4 & 48.1 & 25.6 & 47.4 & 57.4 & 36.1 & 58.4 & 63.3 & 41.3  & 67.1 & 80.2 \\
CenterNet (Multi-Scale) \cite{duan2019centernet}         & Hourglass 104 & 47.0 & 64.5 & 50.7 & 28.9 & 49.9 & 58.9 & 37.5 & 60.3 & 64.8 & 45.1 & 68.3 & 79.7 \\ 
\hline
$x$Net Centers               & Resnet152     & 43.7 & 62.7 & 47.8 & 22.7 & 48.2 & 57.4 & 35.3 & 58.2 & 62.1 & 39.6 & 67.6 & 78.9 \\
$x$Net Centers (Multi-Scale) & Resnet152     & 46.1 & 64.7 & 50.5 & 26.9 & 49.9 & 59.6 & 36.8 & 60.2 & 63.5 & 44.0 & 67.6 & 79.4 \\
$x$Net Corners               & Resnet152     & 45.2 & 64.2 & 49.2 & 25.9 & 48.9 & 57.6 & 36.2 & 59.6 & 62.9 & 42.8 & 67.9 & 78.6 \\
$x$Net Corners (Multi-Scale) & Resnet152     & \textbf{47.8} & \textbf{66.2} & \textbf{52.3} & \textbf{29.7} & \textbf{50.4} & \textbf{60.7} & \textbf{37.8} & \textbf{62.3}  & \textbf{66.0} & \textbf{47.8} & \textbf{69.3} & \textbf{80.8} \\
\hline
\end{tabular}
}
\label{table:comparison_to_others}
\end{table*}

Apart from the mAP numbers and inference times, we studied the difference in performance between the center-based and corner-based object detection based on visual inspection. As shown in Fig. \ref{fig:centervscorner}, there are three main differences that we observed by examining the detection results of the two detectors.
First, the corner-based detector generally produces better detections, while center-based sometimes misses visible objects within the image. Fig. \ref{fig:centervscorner}a shows some examples that demonstrate such difference. Second, the corner-based detector, as shown in Fig. \ref{fig:centervscorner}b, produces refined detections with tighter bounding boxes around the objects compared to the center-based detector.
Finally, the center-based detector performs better when detecting objects that are occluded, while the corner-based detector tends to split the detection into smaller bounding boxes. For example, in the first image of Fig. \ref{fig:centervscorner}c, the bus is occluded by trees, yet the center-based detector was able to detect the bus correctly. On the other hand, the corner-based detector splits the detection into two smaller bounding boxes.

\subsection{Comparison To Other Detectors}

We compared our best detectors for both Centers-$x$Net and Corners-$x$Net to other single stage detectors. We report mAP numbers on MS COCO 'test-dev2017' set. Table \ref{table:comparison_to_others} shows a comprehensive comparison of our top-performing human-crafted architectures when using single and multi-scale input images to the rest of the one-stage detectors. Corners-$x$Net comes on top for both single and multi-scale input images. It also closes the gap between single-stage and two-stage detectors. Centers-$x$Net performs on-par with other anchor-based architectures while only using a single scale per anchor and without using any object-to-anchor assignment optimization. These results demonstrate the effectiveness of using a MatrixNet as a backbone for object detection architectures.

\subsection{Ablation Study}

\subsubsection{MatrixNet Design}
A 5 layer MatrixNet is equivalent to an FPN, so we use that as a baseline for evaluating adding more matrix layers to the backbone. Table \ref{table:ablation_study}a shows the mAP numbers  for different choices of the numbers of the matrix layers. Using 19 layers MatrixNet improves the performance by 5.1 points compared to FPN (5 layers MatrixNet). The extra layers in the 19 layers MatrixNet are much smaller than the FPN layers since each step right or down in the matrix cuts the width or height by half. As a result, the total number of anchors in the 19 layers MatrixNet is ~2.5 times those for FPNs.

\begin{table*}[ht!]
\centering
\footnotesize
\subcaptionbox{The impact of varying the number of matrix layers on the performance of Centers-$x$Net.}{
\begin{tabular}[t]{@{}l|lll@{}}
\textbf{MatrixNet}  & \textbf{$AP$}      & \textbf{$AP_{50}$} & \textbf{$AP_{75}$}  \\
\hline
5 Layers (FPN) & 35.9 & 58.7 & 37.7 \\
11 Layers & 39.2 & 59.0	& 42.7 \\
19 Layers & 41.0 & 60.4 & 45.0 
\end{tabular}
\label{table:fpnvsxnet_table}
}
\quad
\subcaptionbox{The impact of varying the base layer range on the performance of Centers-$x$Net.}{
\begin{tabular}[t]{l|llll}
\textbf{Base Layer Range} & \textbf{$AP$} & \textbf{$AP_{50}$} & \textbf{$AP_{75}$}  \\
\hline
16px-32px & 38.4 & 60.2 & 41.7 \\
24px-48px & 41.0 & 60.4 & 45.0 \\
\multicolumn{3}{l}{}
\end{tabular}
\label{table:base_layer_range}
}
\quad
\subcaptionbox{The impact of varying the training image crop sizes on the performance of Centers-$x$Net.}{
\begin{tabular}[t]{l|lll}
\textbf{Crop Size}  & \textbf{$AP$}      & \textbf{$AP_{50}$} & \textbf{$AP_{75}$}  \\
\hline
512 & 40.3 & 61.4 & 43.5 \\
640 & 41.0 & 60.4 & 45.0  \\
\multicolumn{3}{l}{}
\end{tabular}
\label{table:crop_sizes_centers}
}
\quad
\subcaptionbox{The impact of the backbone on the overall performance of Centers-$x$Net and Corners-$x$Net}{
\begin{tabular}[t]{l|lll||lll|}
  & \multicolumn{3}{c||}{\textbf{Centers}} & \multicolumn{3}{c|}{\textbf{Corners}} \\
\hline
\textbf{Backbone} & \textbf{$AP$}      & \textbf{$AP_{50}$} & \textbf{$AP_{75}$} & \textbf{$AP$}      & \textbf{$AP_{50}$} & \textbf{$AP_{75}$}  \\
\hline
Resnet-50-X &  41.0 & 60.4 & 45.0 & 41.3 & 60.0 & 44.8 \\
Resnet-101-X & 42.3 & 62.1 & 46.4 & 42.3 & 61.0 & 45.8 \\
Resnet-152-X & 43.6 & 62.3 & 47.5 & 44.7 & 63.6 & 48.7 
\end{tabular}
\label{table:backbone_centers_corners_map}
}

\caption{Ablation of design choices on MS-COCO validation set.}
\label{table:ablation_study}

\end{table*}

\begin{figure*}[!h]
\setlength\tabcolsep{1.5pt}
\begin{tabular}{ccccc}
  \includegraphics[height=25mm]{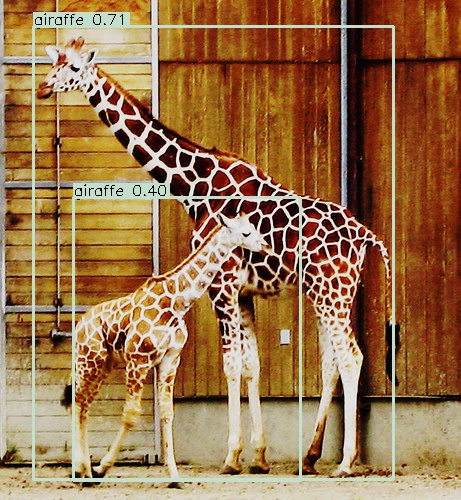} &   \includegraphics[height=25mm]{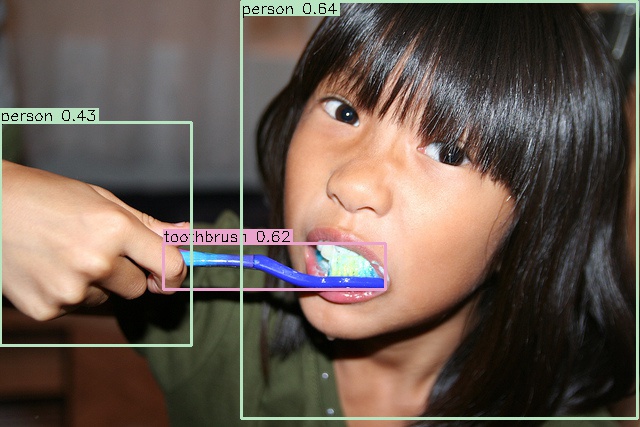} & \includegraphics[height=25mm]{figures/FPN_3.jpg} & \includegraphics[height=25mm]{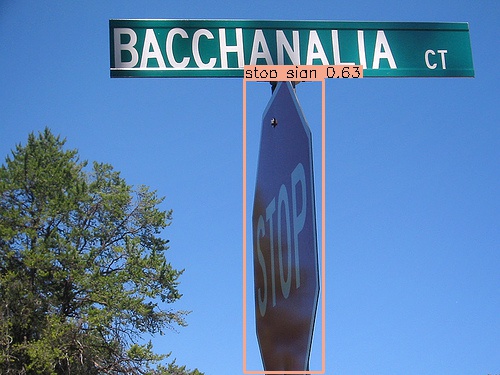} & \includegraphics[height=25mm]{figures/FPN_5.jpg}\\
& & (a) FPN (5 layers MatrixNet).  \\[6pt]
  \includegraphics[height=25mm]{figures/xnet_1.jpg} &   \includegraphics[height=25mm]{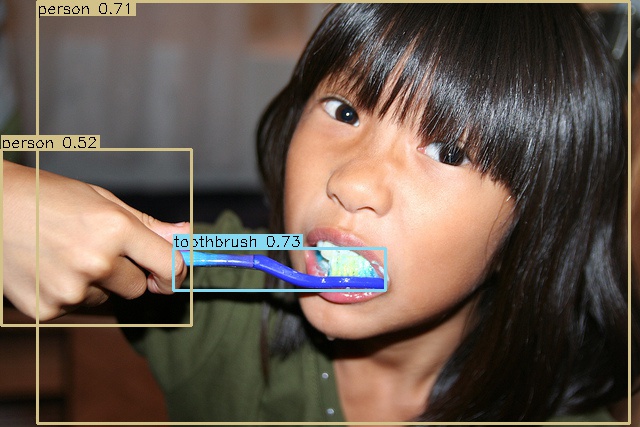} & \includegraphics[height=25mm]{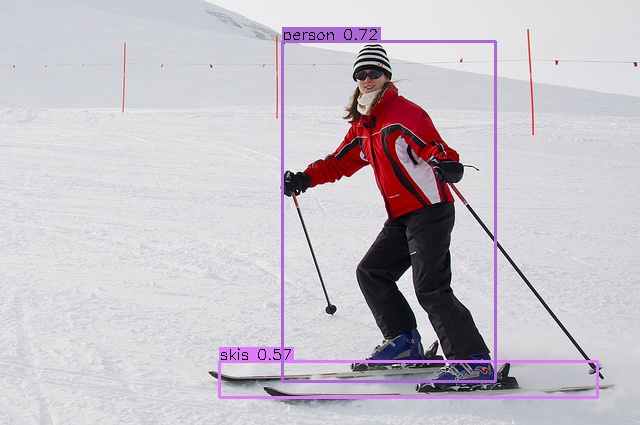} & \includegraphics[height=25mm]{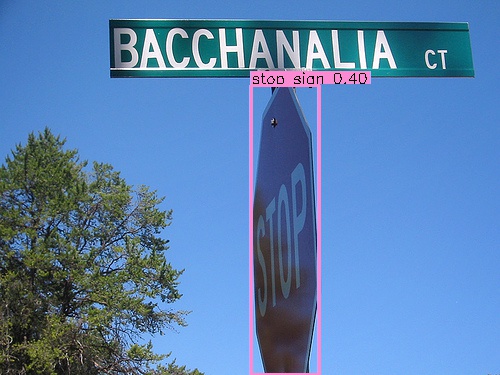} & \includegraphics[height=25mm]{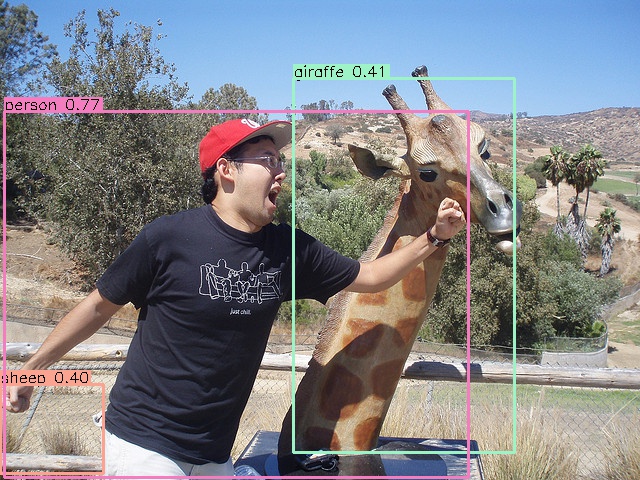}\\
& & (b) 19 layers MatrixNet.  \\[6pt]
\end{tabular}
\vspace{-0.5cm}
\caption{Sample detection results for center-based object detection framework when using FPN (a), and 19 layers MatrixNet (b) as backbones. MatrixNet produces tighter bounding boxes especially for rectangular objects.}
\label{fig:fpnvsmatrixnet}
\end{figure*}


We also did a visual inspection for the detection results for both detectors. Fig.~\ref{fig:fpnvsmatrixnet} shows qualitative examples for the center-based detector when using FPN (5 layers MatrixNet) compared to using 19 layers MatrixNet as backbones. Generally, we observed that using MatrixNet results in better handling of rectangular objects.

\subsubsection{Layer Ranges}

In this experiment, we wanted to examine the effect of the base layer ($l_{1,1}$) range choice on the performance of the detector. We used Centers-$x$Net architecture to evaluate the effect of this hyper-parameter. Table. \ref{table:ablation_study}b shows that using the range of $24$px-$48$px is more effective. The goal for selecting this range is to have a balanced object assignment to all of the matrix layers. Selecting a larger range than $24$px-$48$px (e.g., $32$px-$64$px) would require using a larger training image crops to have enough examples to train the bottom right layers in the matrix. This will require more GPUs and longer training times.
We also found that the choice of layer ranges is as essential for Corners-$x$Net architectures as it is for Centers-$x$Net.


\subsubsection{Training Image Crop Sizes}
During the training, we use scale jitter to scale the image randomly, then we use crops of fixed sizes to train the model. The choice of crop sizes mainly affects the bottom right layers of the MatrixNet. Smaller crop sizes would prevent these layers from having enough objects that span their entire ranges. For Centers-$x$Net, the training crop sizes would impact the performance of the corner regression outputs, and hence the overall performance of the detector. Table \ref{table:ablation_study}c shows the effect of the crop sizes on the overall performance of the Centers-$x$Net architecture. For Corners-$x$Net, the training crop sizes would impact the performance of the center regression outputs. Since center regression output only impacts corner matching, and since we allow for an error of $30\%$, we found that the choice of image crops has little impact on the performance of Corners-$x$Net.


\subsubsection{Backbones}
Backbones act as feature extractors. Hence a better and a larger backbone usually results in a better overall performance for an architecture. Table~\ref{table:ablation_study}d shows the effect of using Resnet50, Resnet101, and Resnet152 on the overall performance of both Centers-$x$Net and Corners-$x$Net architectures.


\section{Conclusion}
\label{sec:Conclusions}
In this work, we introduced MatrixNet, a scale and aspect ratio aware architecture for object detection. We used MatrixNets to solve fundamental limitations of corner based object detection. We also used MatrixNet as a backbone for anchor-based object detection. In both applications, we showed significant improvements in mAP over the baseline.

We view MatrixNet as a backbone that is an improvement over FPN. We demonstrated the impact of using MatrixNet for one-stage object detection, which can be extended in the future to two-stage object detection. MatrixNets can also replace FPNs in other computer vision tasks such as instance segmentation, key-point detection, and panoptic segmentation tasks. Code for the paper is available at \url{https://github.com/arashwan/matrixnet}.

{\small
\bibliographystyle{ieee_fullname}
\bibliography{egpaper}
}

\end{document}